\documentclass[11pt,a4paper]{article}
\usepackage[hyperref]{conll-acl2017}
\usepackage{times}
\usepackage{latexsym}
\usepackage{url}

\usepackage{amsmath}
\usepackage{graphicx}

\DeclareMathOperator*{\SNLI}{\mbox{\sc snli}}
\DeclareMathOperator*{\STS}{\mbox{\sc sts}}

\newcommand{\ignore}[1]{}
\newcommand{\mycomment}[1]{}

\newcommand{\trans}{{}^{\top}}
\newcommand{\by}{{\times}}
\newcommand{\kron}{{\otimes}}
\newcommand{\bove}{Bo{\nolinebreak\hspace{-0.25ex}}VE}
\newcommand{\boves}{Bo{\nolinebreak\hspace{-0.25ex}}VEs}

\aclfinalcopy

\title{ Bag-of-Vector Embeddings of Dependency Graphs for Semantic Induction
}

\author{ Diana Nicoleta Popa \\
  Naver Labs Europe \\
  and Univ.\ Grenoble Alpes \\
  {\tt diana.popa@naverlabs.com} \\\And
  James Henderson\Thanks{~This article reports work done while both authors were at Xerox Research Centre Europe.} \\
  Idiap Research Institute \\ 
  {\tt james.henderson@idiap.ch} 
}

\date{}

\begin{document}
\maketitle
\begin{abstract}

Vector-space models, from word embeddings to neural network parsers, have many
advantages for NLP.  But how to generalise from fixed-length word vectors to a
vector space for arbitrary linguistic structures is still unclear.  In this
paper we propose bag-of-vector embeddings of arbitrary linguistic graphs.  A
bag-of-vector space is the minimal nonparametric extension of 
a vector space, allowing the representation to grow with the size of the
graph, but not tying the representation to any specific tree or graph
structure.  We propose efficient training and inference algorithms based on
tensor factorisation for embedding arbitrary graphs in a bag-of-vector space.
We demonstrate the usefulness of this representation by training bag-of-vector
embeddings of dependency graphs and evaluating them on unsupervised semantic
induction for the Semantic Textual Similarity and Natural Language Inference
tasks.

\end{abstract}

\section{Introduction}

Word embeddings have made a big contribution to recent advances in NLP.  By
representing discrete words in a continuous vector space and by learning
semantically meaningful vectors from distributions in unannotated text, they
are able to capture semantic similarities between words.  But generalising
this success to models of the meaning of phrases and sentences has proven
challenging.  If we continue to use fixed-length vectors to encode arbitrarily
long sentences, then we inevitably lose information as we scale up to
larger semantic structures (e.g.\ \cite{adi2016fine,Blacoe2012,Mitchell-Lapata2010,Socher2011_nips,Kiros2015,le2014distributed,Li-Jurafsky2015,Cho2014,Sutskever2014}).
If we only use vectors to label the nodes of a
traditional linguistic structure, then we lose the advantages of encoding
discrete structures in a continuous space where similarity between structures
can be captured.

In this paper we investigate the minimal extension of a vector space which
allows it to embed arbitrarily large structures, namely a bag-of-vector
space.  These bag-of-vector embeddings (\boves) are nonparametric
representations, meaning that the size of the representation (i.e.\ the number
of vectors in the bag) can grow with the size of the structure that needs to
be embedded.  In our case, we assume that the number of vectors is the same as
the number of nodes in the structure's graph.  But no other information about
the graph has a discrete representation.  All properties and relations in the
graph are encoded in the continuous values of the vectors.

We propose methods for mapping graphs to \bove\ representations and for
learning these mappings.  To take full advantage of the flexibility of the
\bove\ representation, we want these mappings to embed arbitrary graphs.  For
this reason, we propose tensor factorisation algorithms, both for training a
\bove\ model and for inferring a \bove\ given a graph and a model.  Like the
Word2Vec model of word embeddings \cite{word2vec2_nips}, tensor factorisation
uses a reconstruction loss, where each observed relation is predicted
independently conditioned on the latent vector representation.  This
conditional independence allows tensor factorisation to model arbitrary
graphs.  As well as stochastic gradient descent, we propose efficient
alternating least squares algorithms for optimising this reconstruction loss,
both for training models and for inferring \boves\ for a new graph.

\begin{figure}[tb]
  \centerline{\includegraphics[scale=0.5]{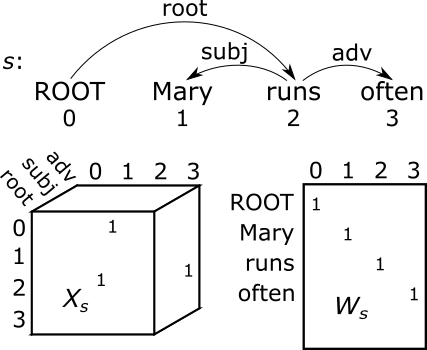}}
  \caption{A dependency parse graph $s$ and its relation tensor $X_s$ and
    property matrix $W_s$, where all cells not shown as 1 are 0.}
  \label{fig:parse}
\end{figure}

As an example of the usefulness of these algorithms, we learn \bove\ models
for embedding dependency parses of sentences, such as $s$ in
Figure~\ref{fig:parse}.  In these embeddings, each
vector corresponds to a word token in the sentence.  We can think of these
token vectors as context-dependent word embeddings.  The \bove\ model learns 
to embed dependency relations by adding features to these token vectors which
specify the features of the tokens it is related to.  So, each token vector
encodes information about its context in the graph, as well as information
about its word.

We evaluate this property by comparing our \bove\ embeddings
to bags of Word2Vec word embeddings.  Initialising a \bove\ model with these
same word embeddings, we train a model of relations and infer
context-dependent vectors for each word token.  We then evaluate these two
bag-of-vector representations in unsupervised models of two sentence-level
semantic tasks, Semantic Textual Similarity (STS) and the Stanford Natural
Language Inference (SNLI) dataset.  Results show that training a \bove\ model
does extract semantically meaningful information from the distributions in a
corpus of syntactically-parsed text.

In the rest of this paper, we define our tensor factorisation algorithm for
learning the parameters of a model, and define our inference algorithm for
computing an embedding for a graph given these parameters.
We evaluate these algorithms on there ability to extend word embeddings, using
two tasks which demonstrate that the resulting bag-of-vector embeddings induce
a semantic representation which is more informative than word embeddings.

\section{Bag-of-Vector Embeddings}

Our proposed method for embedding graphs in a bag-of-vector space has two
steps: first we learn the parameters of a model from a large corpus of example
graphs, then we compute the embedding of a new graph using those parameters.
The training phase uses correlations in the data to find good compressions
into the limited capacity of each vector.  It is the encoding of these
correlations in the final embeddings which makes these representations
semantically meaningful. 

\begin{figure*}[tb]
  (a)
  \begin{tabular}{l}
    \includegraphics[scale=0.55]{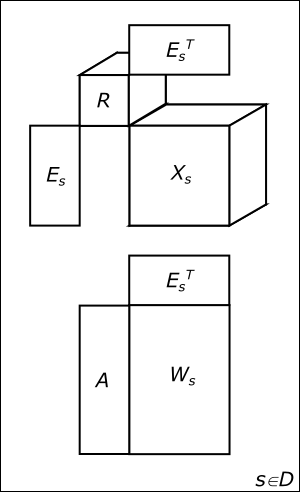}
  \end{tabular}
  ~~~~
  \begin{tabular}{l}
    (b)
  \begin{tabular}{ll}
    \(D\) 
    & corpus of sentence graphs
    \\
    \(W_{s}: c\by {|s|}\) 
    & predicates of tokens in $s{\in}D$
    \\
    \(X_{s}: d\by {|s|}\by {|s|}\)
    & relations between tokens in $s{\in}D$
    \\[1ex]
    \(P: c\by r\) 
    & predicate vector embeddings
    \\
    \(R: d\by r\by r\) 
    & relation matrix embeddings
    \\
    \(E_{s}: {|s|}\by r\) 
    & token vector embeddings of tokens in $s{\in}D$
    \\[1ex]
    $r$ & size of the embedding vectors \\
    $c$ & number of unary predicates  \\
    $d$ & number of binary relations  \\
    $|s|$ & number of tokens in sentence $s$
  \end{tabular} 
  \\ ~ \\ ~ \\
  (c) ~~~\(\displaystyle \forall s, \)
  \begin{tabular}{l}
    \(\displaystyle X_s \approx E_s \cdot R \cdot E_s\!\trans \)
    \\
    \(\displaystyle W_s \approx P \cdot E_s\!\trans \)
    \\ %~ \\ 
  \end{tabular}
  \end{tabular}
  \caption{(a) A depiction of the tensor decomposition model, 
    with the plate
    indicating one such pair per sentence in the data, with $R$ and $P$ shared
    across sentences.  (b) The symbols used in the definition of the model.  
    (c) The tensor decomposition in equations.}
  \label{fig:tensor}
\end{figure*}

We propose tensor factorisation algorithms for both learning a \bove\ model
and inferring \boves\ given a model.  In tensor factorisation, the graph is
encoded as a tensor for relations and a matrix for properties, as illustrated
in Figure~\ref{fig:parse}.  The relations
in the graph are encoded as an entity-by-label-by-entity tensor $X_s$ of indicator
variables, where entities are the nodes of the graph and a 1 indicates a
relation with the given label between the given entities.  All other cells in
the tensor are zero.  The properties of nodes in the graph are encoded as an
entity-by-label matrix $W_s$ of indicator variables, where a 1 indicates a
property with the given label for the given entity.  As illustrated in
Figure~\ref{fig:tensor}a, the embedding
learns to reconstruct each cell of the matrix as the dot product between the
vector for the entity and the vector for the property, and learns to
reconstruct each cell of the tensor as the dot product between the vectors for
the two entities and a matrix for the relation.  In this work we assume
squared loss for each of these reconstruction predictions, because it leads to
efficient inference procedures.

Training the tensor factorisation model results in vectors for all the
properties, plus matrices for all the relations.  In our setting the
properties are words and part-of-speech (PoS) tags, and the relations are
syntactic dependencies and string adjacency.  We refer to these as the type
embeddings.  At test time, given an input graph, we freeze these type
embeddings and infer one vector for each node in the graph.  In our setting
the nodes are tokens of words in the sentence.  These are the token vectors.
The bag of token vectors is the embedding of the graph.

\subsection{The \bove\ Model}

Our use of tensor factorisation is closely related to the RESCAL model
\cite{nickel2011,nickel2012}.  RESCAL is designed as a model of large
relational databases, with large numbers of entities, properties of entities
and (binary) relations between entities.
Unlike in the RESCAL setting, the entities in our datasets are partitioned
into sentences.  No relations are possible between tokens in different
sentences, so representing a corpus as one big database where anything can be
related to anything is not appropriate.  Also unlike in RESCAL, our target use
case is transductive, not abductive; we want to train the parameters of a
model which can then be applied to compute the embedding of a previously
unseen parsed sentence.  And we want the inference necessary at test time to
be fast, which excludes the possibility of re-factorising the training set.

Given a set
of sentence graphs $D$, we define the tensor dimension sizes $r$, $c$, $d$
and $|s|$ as in Figure~\ref{fig:tensor}b.  The data $D$ is encoded in
indicator tensors $X_s,W_s$, also specified in Figure~\ref{fig:tensor}b.  For each
sentence graph $s{\in}D$, there is a 2-dimensional tensor $W_{s}$ that indicates
which unary predicates label which tokens, and a 3-dimensional tensor $X_{s}$
that indicates which binary relations exist between two tokens.  We want to
factorise these tensors into the real-valued embedding tensors $P$ for unary
predicates, $R$ for binary relations, and $E_{s}$ for the tokens in each
sentence $s{\in}D$, also defined in Figure~\ref{fig:tensor}b.

This tensor factorisation is depicted in Figure~\ref{fig:tensor}a.  The
objective of the factorisation is to be able to reconstruct $W_s$ and $X_s$
from $P$, $R$ and $E_s$, for all $s{\in}D$:
\vspace{-0.5ex}
\begin{align}
W_s \approx~& P \cdot E_s\!\trans 
\label{eqn:W}\\
X_s \approx~& E_s \cdot R \cdot E_s\!\trans 
\label{eqn:X}
\\[-4ex]\nonumber
\end{align}
We use quadratic loss for efficiency reasons.  Adding quadratic ($L_2$)
regularisation, we get the objective function:
\vspace{-0.5ex}
\begin{align}
\label{eqn:objective}
&
\sum_{s{\in}D}  
 || W_{s} - P \cdot E_{s}\!\trans ||^2_2 
\\&
+\sum_{s{\in}D} 
 \alpha || X_{s} - E_{s} \cdot R \cdot E_{s}\!\trans ||^2_2
\nonumber\\&
+ \lambda_P ||P_{ij}||^2_2
~~+ \lambda_R ||R_{ijk}||^2_2
+\sum_{s{\in}D} \lambda_E ||E_{sij}||^2_2
\nonumber
\\[-4ex]\nonumber
\end{align}
where $\alpha$ is a hyperparameters determining the relative importance of
embedding relations versus embedding properties, and $\lambda_P$, $\lambda_R$
and $\lambda_E$ are hyperparameters that determine the strength of
regularisation.  

In addition to $L_2$ regularisation, we have also tried $L_1$ regularisation and
nuclear-norm regularisation for $R$ and $P$.  To do nuclear-norm
regularisation on $R$, we run SVD on each relation's slice of $R$, apply $L_1$
regularisation on the resulting eigenvalues, and then reconstruct the slice of
$R$ with the reduced eigenvalues.  This in effect regularises the rank of the
matrix for each relation, which is particularly motivated because predicting
labelled relations is relatively easy.  There are a very large number of
words, so the full rank of the model is needed to predict words.  But most
relations are much easier to predict, and thus need a much smaller rank.

Training our model optimises the above objective to find values for $P,R$, and
values for $E_s$ for all training graphs $W_s,X_s$ in the training corpus $D$.  At
testing time we are given a new graph $W_s,X_s$ and we optimise a new $E_s$
while keeping $P,R$ fixed.  The rows of the matrix $E_s$ are the vector
representations of the entities in $W_s,X_s$.  The rows of $E_s$ are
exchangeable, which is why we refer to them as a bag-of-vectors.

\subsection{Training the \bove\ Model}
\label{sec:training}

Given the objective in equation~\ref{eqn:objective}, it is straightforward to
define a stochastic gradient descent (SGD) algorithm for training a
\bove\ model.  In this section, we focus on an alternating least squares (ALS)
optimisation algorithms for training \bove\ models.  Provided that the size
$r$ of the vectors is not too large, this algorithm is much faster than SGD.

This ALS algorithm is inspired by the RESCAL algorithm
\cite{nickel2011,nickel2012}.  Like the RESCAL algorithm, we cycle between
updating $P$ and $R$ given $E$, and updating $E$ given $P$ and $R$.  For $P$
and $R$, there is a closed form solution to find the exact minimum of the
objective.  To find the optimal $E$ given $P$ and $R$, we need to iteratively
compute the least-squares solution for a new value of $E$ given the old value
of $E$.  We do a few steps of this iteration for each update of $P$ and $R$,
as discussed below.  First we give details of the closed form solution for
updating $P$ and $R$.

To update $P$, we construct one matrix $W$ whose rows are the concatenation
($cat^{row}$) of the rows from the matrices $W_s$ for all the sentences
$s{\in}D$.
\vspace{-0.5ex}
\begin{align*}
W =~& cat^{row}_{s{\in}D}(W_s)
\\[-4ex]\nonumber
\end{align*}
So $W$ has $c$ rows and as many columns as there are tokens in the training
corpus.  Similarly, we construct a matrix $E$, which has $r$ columns and as
many rows as tokens in the corpus.
\vspace{-0.5ex}
\begin{align*}
E =~& cat^{col}_{s{\in}D}(E_s)
\\[-4ex]\nonumber
\end{align*}
We then find the optimal $P$ to minimise the regularised quadratic loss for
\vspace{-0.5ex}
\begin{align*}
W \approx~& P \cdot E\/\trans
\\[-4ex]\nonumber
\end{align*}
Using standard linear algebra, this gives us:
\vspace{-0.5ex}
\begin{align*}
P =~& W \cdot E \cdot (E\trans\! \cdot E + \lambda_P I\/)^{-1}
\\[-4ex]\nonumber
\end{align*}
Note that this requires computing the inverse of an $r\by r$ matrix.

To update $R$, we first represent $R$ as a matrix $R^\prime$ by listing each
slice of the matrix $R_i$ in a row vector by concatenating its rows
($vec^{row}$).
\vspace{-0.5ex}
\begin{align*}
R^\prime_i =~& vec^{row}(R_i)
\\[-4ex]\nonumber
\end{align*}
So $R^\prime$ has $d$ rows and $r^2$ columns.  
Secondly, we map each sentence's tensor into a matrix $X^{\prime}_{s}$ by
enumerating pairs of tokens for each column of $X^{\prime}_{s}$, so each
matrix $X_{si}$ becomes row $i$ of $X^{\prime}_{s}$.
\vspace{-0.5ex}
\begin{align*}
X^{\prime}_{si} =~& vec^{row}(X_{si})
\\[-4ex]\nonumber
\end{align*}
Thirdly, we map each sentence's embedding matrix into a larger matrix
$E^{\prime}_{s}$ with $r^2$ columns and a row for every pair of tokens in the
sentence.  If  $i,j$ is the $p^{th}$ pair, then the $p^{th}$ row of
$E^\prime_{s}$ is the vectorisation of the outer product between the embedding
of $i$ and the embedding of $j$.  We can write this formally using the
Kronecker product $\kron$:
\vspace{-0.5ex}
\begin{align*}
E^{\prime}_{s} =~& E_{s}\kron E_{s}
\\[-4ex]\nonumber
\end{align*}

We then concatenate the matrices for the different sentences as done above for
updating $P$, to get $X^{\prime}$ and $E^{\prime}$.  
\vspace{-0.5ex}
\begin{align*}
X^\prime =~& cat^{row}_{s{\in}D}(X^\prime_s)
\\
E^\prime =~& cat^{col}_{s{\in}D}(E^\prime_s)
\\[-4ex]\nonumber
\end{align*}

Based on the equation 
$vec^{row}(A\cdot B\cdot C){=}vec^{row}(B)\cdot (A\kron B)$, finding the
optimal $R$ can then be done by finding the optimal $R^\prime$ to minimise the
regularised quadratic loss for 
\vspace{-0.5ex}
\begin{align*}
X^{\prime} \approx~& R^\prime \cdot E^{\prime}\/\trans
\\[-4ex]\nonumber
\end{align*}
Which gives us:
\vspace{-0.5ex}
\begin{align*}
R^\prime =~& X^{\prime} \cdot E^{\prime} \cdot (E^{\prime}\/\trans\! \cdot E^{\prime} + \lambda_R I\/)^{-1}
\\[-4ex]\nonumber
\end{align*}
Note that this least-squares problem requires finding the inverse of an
$r^2\by r^2$ matrix, which becomes very expensive as $r$ becomes large.  Our
experiments so far have been done with ranks where this is not the limiting
factor ($r\leq 100$).

To update $E$, we take advantage of the fact that the tokens in different
sentences cannot be in relations with each other.  This allows us to optimise
the embeddings $E_s$ for all sentences $s$ independently of each other.
Updating each $E_s$ is an instance of the same problem as updating the entity
embeddings in RESCAL, only for a smaller relation tensor and property matrix.
We use a modified version of the update procedure of RESCAL, applying it once
for each 
sentence.  The RESCAL procedure updates $E_s$ using the old version of $E_s$
for one side of the relation tensor.  This gives three sets of equations which
need to be optimised, one for each occurrence of $E_s$ in Equations~\ref{eqn:W}
and~\ref{eqn:X}.
\vspace{-0.5ex}
\begin{align}
\nonumber
W_s\!\trans \approx~& E^{t}_s \cdot P\trans 
\\ \nonumber
X_s \approx~& E^{t}_s \cdot R \cdot E^{(t-1)}_s\!\trans 
\\ \nonumber
X_s\!\trans \approx~& E^{t}_s \cdot R\trans\! \cdot E^{(t-1)}_s\!\trans 
\\[-4ex]\nonumber
\end{align}
where $E^{(t-1)}_s$ is the old version and $E^{t}_s$ is the new matrix we want
to optimise.  Adding our $\alpha$ weighting to the equations for the RESCAL
procedure we get the following solution to the regularised least squares
objective:
\vspace{-0.5ex}
\begin{align}
  &
  E^{t}_s =
  \label{eqn:updateE}
  \\&
  ~ cat^{row}(W_s\!\trans\!,~ \alpha X_s,~ \alpha X_s\!\trans)
  \cdot F\trans\! \cdot (F \!\cdot\! F\trans\! + \lambda_E I\/)^{-1}
  \nonumber
\\[-4ex]\nonumber
\end{align}
where
\vspace{-0.5ex}
\begin{align*}
F =~& cat^{row}(P\trans\!,~ \alpha R \cdot E^{(t-1)}_s\!\trans\!,~ \alpha R\trans \cdot E^{(t-1)}_s\!\trans)
\\[-4ex]\nonumber
\end{align*}
and $cat^{row}(A,\,B,\,C)$ concatenates the rows of $A$, $B$ and $C$.

However, we found that with this procedure, the new values $E^{t}_s$
tend to overcompensate for the errors in the old values $E^{(t-1)}_s$,
resulting in oscillation.  This may be because, in our setting, the vectors
for $E_s$ are much bigger than they need to be to represent $X_s$; most of
that capacity is only needed for representing $W_s$.  We addressed this
problem by running two consecutive iterations of this optimisation, keeping
$R$ and $P$ fixed, to get $E^{t}_s$ and $E^{t+1}_s$.  As our new value of
$E_s$, we used the average of $E^{t}_s$ and $E^{t+1}_s$:
\vspace{-0.5ex}
\begin{align}
E_s =~& \frac{E^{t}_s + E^{t+1}_s}{2}
\label{eqn:aveE}
\\[-4ex]\nonumber
\end{align}
This average (in $E_s \cdot R \cdot E_s\!\trans$) is a better approximation to
the combined effect (in $E^{t}_s \cdot R \cdot E^{(t+1)}_s\!\trans$) of the
$t$ and $t+1$ matrices than using the $t+1$ matrix alone (as in RESCAL).
This average can be seen as a two-step process which first infers $E^{t}_s$
and $E^{t+1}_s$ separately and then projects into a sub-space where they are
equal.  We run one such average-of-two update of all $E_s$ in $E$ in between
each update of $R$ and $P$.

Overall, our objective in this alternating least squares training is only to
find good values of $R$ and $P$.  As discussed in the next subsection, at test
time we are given a new sentence's $X_s$ and $W_s$, and we want to compute
$E_s$ for this new sentence, keeping $R$ and $P$ fixed to the values learned
during training.  To make the training setup as similar as possible to the
testing setup, in both cases we initialise the model with zero values for $E$.
In training, we initialise $P$ with random values and $R$ with zeros.  We also
periodically re-initialise $E$ to zeros and run several iterations of updating
$E$ (as in testing) before returning to the alternating least squares
optimisation described above.  We stop training when the squared loss stops
improving by more than 0.1\% at each iteration.

\subsection{Inference of a \bove\ for a Graph}
\label{sec:inference}

Because we are assuming a transductive learning setting, at test time we are
given a new sentence and its $X_s$ and $W_s$, and we want to compute $E_s$ for
this new sentence, keeping $R$ and $P$ fixed to the values learned during
training.  The objective function for this inference of $E_s$ remains the same
as in training (equation~\ref{eqn:objective}), so again this optimisation can
be done with either SGD or ALS.  But because there is no need to optimise $R$,
there is no need to compute the inverse of an $r^2\by r^2$ matrix, making ALS
faster even for larger embedding sizes $r$.

We initialise $E^{t=0}_s$ to zero values, and run several iterations of the
ALS optimisation procedure described in section~\ref{sec:training}, using
equations~\ref{eqn:updateE} and~\ref{eqn:aveE}.  During the first update,
$E^{t=1}_s$ is not effected by $R$ and $X_s$, so each entity only receives
features from its properties in $P$.
\vspace{-0.5ex}
\begin{align*}
E^{t=1}_s =~& W_s\!\trans\! \cdot P \cdot (P\trans\! \cdot P + \lambda_E I\/)^{-1}
\\[-4ex]\nonumber
\end{align*}
In the second update, these features are combined with features propagated
from the entity's immediately related entities.  With each update, features
from farther away in the graph have an impact on the entity's embedding.
After the first update, we apply the averaging procedure described in
section~\ref{sec:training} to every update, applying equation~\ref{eqn:aveE}
with the previous average as $E^{t}_s$ and the new result of
equation~\ref{eqn:updateE} as $E^{t+1}_s$.

While the number of iterations performed during testing could be
determined in a number of ways, in our experiments we simply use a
fixed number of iterations (30).

\section{Related Work}

As outlined above, the proposed model is closely related to RESCAL,
which was developed for learning embeddings for entities in a large
relational database.  Our model differs in that it learns from many
small graphs, rather than one big one, and it is targeted at computing
embeddings for new entities not in the training set.  \citet{riedel2013}
combine entities from a knowledge
base with entities in text and jointly factorises them.  But they do
not use tensor factorisation methods, and the above two differences
also apply.

Neural network (NN) models have been used to compute embeddings for
parsed sentences.  They either use a fixed-length vector for an arbitrarily long
sentence (e.g.\ \cite{Socher2011_nips}), or they keep the
original sentence structure and decorate it with vectors
(e.g.\ \cite{Henderson03_naacl,socher13}).  Ours is the first non-parametric
vector-based model that does not keep the entire structure.  However,
attention-based NN models could be used in this way.  For example in machine
translation, \citet{bahdanau14} take a source sequence of words and encode it
as a sequence of vectors, which are then used in an attention-based NN model
to generate the target sequence of words for the translation.  Keeping the
original ordering of words is not fundamental to this method, and thus it
could be interpreted as a bag-of-vector embedding method for sequences.
However, it is not at all clear how to generalise such a sequence-encoding
method to 
arbitrary graphs.  Similarly, Kalman filters have been used to induce vector
representations of word tokens in their sequential context
\cite{Belanger2015}, but it is not clear how to generalise this to arbitrary
graphs.

Previous work on context-sensitive word embeddings has typically treated this
as a word sense disambiguation problem, with one vector per word sense
(e.g.\ \cite{Neelakantan2015,Liu2015_ijcai}).
In contrast, for our method every different context results in a different
word token vector.

\section{Empirical Evaluation}

As an example of the usefulness of the proposed algorithms and
\bove\ representations, we evaluate them for unsupervised semantic induction.
We train \bove\ models for embedding dependency parses of sentences, and use
the resulting \boves\ in unsupervised models of Semantic Textual Similarity
(STS) \cite{agirre2014semeval,Agirre2015SemEval2015T2} and the Stanford
Natural Language 
Inference (SNLI) dataset \cite{snli:emnlp2015}.  Training is done on a
standard treebank, without looking at the data for the task.  Then the
sentences for the task data are parsed and the \boves\ for these parses are
inferred.  Then the semantic relationship between two sentences is predicted
using an alignment between the elements in the two \boves.

In \boves\ of dependency parses, each vector corresponds to a word token in
the sentence, and encodes both the features of that word and its context.  As
a strong baseline, we use Word2Vec embeddings as representations that just
encode features of the word, without its context in the sentence.  This
baseline is a good representative of the state-of-the-art in unsupervised
semantic induction.  Thus, we compare two bag-of-vector representations, one a
bag of word-embeddings, and another a bag-of-vector embedding of the parsed
sentence.

To provide a direct comparison to this word-embeddings baseline, we initialise
the word type embeddings in our \bove\ model to Word2Vec word embeddings.
This also has the advantage that the model can leverage the fact that these
embeddings have been trained on a very large corpus.  The word type vectors
are then frozen, and the other type embeddings (PoS vectors, dependency
matrices and an adjacency matrix) are trained to optimise the regularised reconstruction loss in
equation~\ref{eqn:objective}, using the CoNLL 2009 syntactic dependency
corpus.  These trained parameters are then also frozen and used to infer
\boves\ of the test sentences.

Given two bag-of-vector representations for two sentences, we use an
alignment-based model to predict either similarity (for STS) or entailment
(for SNLI) between the two sentences.  In both cases, the score for the pair
is the score of the best alignment between the pair of bags, and the
evaluation measure is a function of the ranked list of these scores.

\subsection{Experimental Setup}

\paragraph{Training corpus}
As the training corpus,
we use the CoNLL 2009 \cite{Hajivc09_conll-st} syntactic dependency parses,
derived from the Penn Treebank (PTB) collection of parsed and PoS-tagged Wall
Street Journal texts.  It consists of 40k parsed
sentences, 69 unique syntactic dependencies, 20k unique word types and 1
million word tokens, with an average of 25 word tokens per sentence.  We also
add 1 adjacency relation to the parse graph.  We impose a frequency
threshold of 2 for word types and PoS tags and 1000 for syntactic
dependencies. Word types that appear with frequency lower than the threshold
are replaced by an 'UNKNOWN\_$\langle$POS$\rangle$' tag, where
$\langle$POS$\rangle$ stands for the part of speech tag associated to the
word. Similarly all infrequent PoS tags are replaced by the tag
'UNKNOWN\_POSTAG' and all infrequent relations are replaced by an
'UNKNOWN\_RELATION' tag. We use a generic NB tag to replace all numbers and a
PUNCT tag for all punctuation signs.
For each sentence, we populate a matrix
and a tensor of indicator variables with the corresponding information.

\paragraph{Training the {\bove} model}
At training time, we fix the word type embeddings to their corresponding
pre-trained GoogleNews Word2Vec values whenever
available.\footnote{\url{https://code.google.com/archive/p/word2vec/}} These
vectors were trained with the Word2Vec software \cite{word2vec2_nips} applied
to about 100 billion words of the Google-News dataset, and have 300
dimensions.  Then we train parameters for the remaining word types, PoS
tags and binary relations. Although it is possible to learn word type
representations from scratch, setting the word types to their Word2Vec
vectors allows us to leverage word embeddings trained on a larger
corpus. However, as will be seen below, the PTB corpus is sufficiently
large for efficiently learning embeddings for PoS tags and relations. Training
the model is done using mini-batch stochastic gradient descent with Adagrad \cite{adagrad2011}
and contrastive negative sampling to minimise the objective
function in equation~\ref{eqn:objective}.
Because the embedding size $r$ is 300, the ALS algorithm would be slow due the
computation of $r^2\by r^2$ matrix inverses.

\paragraph{Inferring the {\bove} of a sentence }
At inference time, we obtain a \bove\ for each
sentence, given fixed word type and relation embeddings learned during the
training phase. All sentences are tokenised, PoS tagged and parsed 
and then token embeddings are inferred using the alternating
least squares (ALS) method from Sections~\ref{sec:training} and~\ref{sec:inference}.

\subsection{Scores for Pairs of \boves}

Both STS and SNLI are tasks where we need to score pairs of sentences.  After
inferring a \bove\ for each of the sentences, we predict the semantic
relationship between the sentences by computing a score based on an alignment
between the elements in the two {\bove}s.  

For SNLI, we want the score to reflect how well a sentence $S_{2}$ is entailed
by a sentence $S_{1}$, so we use an asymmetric alignment where all the words in
$S_{2}$ are aligned to some word in $S_{1}$:
\vspace{-0.5ex}
\begin{align*}
&\sum_{j=0}^{|S_{2}|} \max_{0 \leq i \leq |S_{1}|} \cos(S_{1i}, S_{2j})
\\[-4ex]\nonumber
\end{align*}
where $\cos$ is the cosine between the two vectors.  
This score has the property that it grows with the length of $S_{2}$, since
each component of the sum is positive.  In the SNLI data, we observe a
negative correlation between the entailed sentence size and the entailment
score, indicating that a pair of sentences is more likely to be considered in
an entailment relation if the second sentence is shorter.  This
correlation seems to be an idiosyncrasy of the dataset, and because we are only
considering unsupervised models, modelling this correlation should
not be allowed.  For this reason, we only consider scores which are
independent of sentence length.  To make the cosine score independent of
sentence length, we divide 
the entailment score between the 2 sentences by the length of the entailed
sentence and use this as our scoring function for evaluations.
\vspace{-0.5ex}
\begin{align}
  \SNLI(S_{1}, S_{2}) =~&
  \frac{\displaystyle \sum_{j=0}^{|S_{2}|} \max_{0 \leq i \leq |S_{1}|} \cos(S_{1i}, S_{2j})}{|S_{2}|}
  \label{eqn:scoreSNLI}
\\[-4ex]\nonumber
\end{align}
This means that, for each word of the entailed sentence, we find a word
in the entailing sentence that best entails it, and then average over these
alignment scores.\footnote{Note that these maximums over individual alignments
  are also global maximums, since the individual alignments are independent.}
This gives us an indication of how well one sentence is entailed by the other.

For STS, we want the score to reflect the similarity between $S_{2}$ and
$S_{1}$, which is a symmetric relation.  As is typically done with
alignment-based measures, to get a symmetric score, we compute the asymmetric
score in equation~\ref{eqn:scoreSNLI} in both directions, and then use the harmonic mean
between these two scores.
\vspace{-0.5ex}
\begin{align}
  \STS(S_{1}, S_{2}) =\,&
  \frac{ 2 \SNLI(S_{1}, S_{2}) \SNLI(S_{2}, S_{1}) }{
    \SNLI(S_{1}, S_{2}) + \SNLI(S_{2}, S_{1}) }
  \label{eqn:scoreSTS}
\\[-4ex]\nonumber
\end{align}

\subsection{Semantic Textual Similarity}

Semantic Textual Similarity (STS) \cite{agirre2014semeval,Agirre2015SemEval2015T2} is a shared task for systems designed to
measure the degree of semantic equivalence of pairs of texts. Most submissions
to the STS tasks use supervised models that are trained and tuned on the
provided training data or on similar datasets from earlier versions of the
task, and many use additional knowledge resources
(e.g.\ \cite{sultan2015dls}).  We use this data in a 
fully unsupervised setting, where our only external resources are corpora of
raw of parsed text.  We use the STS-2014 data as
a development set to evaluate {\bove}s trained with a few different
hyperparameters settings.  The best results are reported in the top half of
Table~\ref{tab:results-STS}.  The STS-2014 dataset consists of 6 subsets
covering different domains, which we report separately, along with their
average score.  We then evaluated this one best model on the STS-2015 data.
Results are shown in the bottom half of Table~\ref{tab:results-STS}.  The
STS-2015 dataset consists of 5 subsets covering different domains:
answers-forums (Q\&A in public forums), answers-students, belief, headlines
(news headlines) and images (image captions).  We report the standard
evaluation measure provided within the SemEval-2015 Task 2 \cite{Agirre2015SemEval2015T2}
based on the mean Pearson correlation between
the gold scores and the predicted scores.

\begin{table}[htb]
\begin{tabular}{|l@{~}|@{~}c@{~}||@{~}c@{~}|c|}
\hline
Corpus & pairs & Bag of W2V & \bove \\
\hline\hline
{deft-forum 2014} & 450 & {0.4334} & {\bf 0.4386}\\
\hline
{deft-news 2014} & 300 & {0.6583} & {\bf0.6795}\\
\hline
{images 2014} & 750 & {0.7398} & {\bf 0.7410}\\
\hline
{headlines 2014} & 750 & {0.6166} & {\bf 0.6301}\\
\hline
{OnWN 2014} & 750 & {\bf 0.6857} & {0.6800}\\
\hline
{tweet-news 2014} & 750 & {\bf 0.7235} & {0.7187}\\
\hline
{\bf 2014 mean} &  & {0.6429} & {\bf 0.6480}\\
\hline\hline
{answers-forums} & 375 & {0.6133} &{\textbf{0.6174}}\\
\hline
{answers-students} & 750 & {0.7123}&{\textbf{0.7145}}\\
\hline
{belief} & 375 & {\textbf{0.7384}} &{0.7295} \\
\hline
{headlines} & 750 & {0.6904} &{\textbf{0.7033}} \\
\hline
{images} & 750 & {0.8008} &{\textbf{0.8073}} \\
\hline
{\bf 2015 mean} &  & {0.7110} &{\textbf{0.7144}} \\
\hline
\end{tabular}
\caption{ Pearson correlations between the gold STS scores and the scores
  $\STS(S_1,S_2)$ between the \boves\ for the two sentences.  }
\label{tab:results-STS}
\end{table}

As can be seen in Table~\ref{tab:results-STS},
the \bove\ model shows an improvement over the
Word2Vec model in four out of six datasets for STS-2014 and in four out of
five datasets for STS-2015, and in both cases in the average across datasets.
These differences are not statistically significant.

\subsection{Natural Language Inference}

The Stanford Natural Language Inference (SNLI) corpus \cite{snli:emnlp2015}
is a benchmark
for the evaluation of systems on the task of textual entailment. It consists
of 570k human-written English sentence pairs labelled as entailment,
contradiction or neutral and divided into pre-defined train, development and
test sets. While most approaches using SNLI keep the 3-class structure, we
focus on detecting whether two sentences are in an entailment relation or not
and thus combine the 'neutral' and 'contradiction' labels under one single
'non-entailment' label. 

For these evaluations, we used the same \bove\ model which gave the best
results on the development set for STS, and ran evaluations on both
the SNLI development set and the SNLI test sets.  We infer the {\boves} for
the sentences in the SNLI data using this model, and then each pair of
sentences is assigned a score using equation~\ref{eqn:scoreSNLI}.  These
scores are ranked, and we report results in terms of average precision of
these rankings.  The results are shown in Table~\ref{tab:results-SNLI}.

\begin{table}[htb]
\begin{tabular}{|l||c|c|c|}
\hline
Corpus  & pairs & Bag of W2V & \bove \\
\hline\hline
SNLI-dev & {10000} & {64.47\%} & {\bf 65.74\%}\\
\hline
SNLI-test & {10000} & {63.04\%} & {\bf 64.01\%}\\
\hline
\end{tabular}
\caption{
Average precision on SNLI data of lists ranked by the score $\SNLI(S_1,S_2)$
between the \boves\ for the two sentences. 
}
\label{tab:results-SNLI}
\end{table}

As can be seen in Table~\ref{tab:results-SNLI}, the \bove\ model shows an
improvement over the Word2Vec model on both datasets.  These differences are
statistically significant.

\section{Conclusions}

This paper proposes methods for training and inferring bag-of-vector
embeddings of linguistic graphs.
The above empirical results indicate that these \bove\ models succeed in
inducing semantic information from a corpus of parsed text.  In particular,
the way the \bove\ model embeds information about the syntactic
context of a word token results in a better measure of semantic similarity
than using word embeddings, as reflected in better unsupervised models of
Semantic Textual Similarity and the Stanford Natural Language Inference
dataset.

In addition, several theoretical properties motivate the proposed algorithms
for learning a model of embedding graphs in a bag-of-vectors and for inferring
the bag-of-vector embedding of a graph given such a model.  The use of
bag-of-vector spaces as the representation eliminates the need to maintain
discrete relationships as part of the representation, but still allows the
embedding of arbitrarily large graphs in arbitrarily large (nonparametric)
representations.  The use of reconstruction loss as the objective allows these
methods to be applied to arbitrary graphs.  The alternating-least-squares
algorithms scale well to large datasets and make inference at test time
efficient.

Future work includes the use of the trained \bove\ representations in
supervised semantic tasks.  In this context, \boves\ are a natural match with
attention shifting neural network models, where their content-based access to
vectors in the bag eliminates the need for other data structures, such as a
stack or tape.  This approach should allow many NLP tasks which traditionally
rely on discrete structured representations to take advantage of the
continuous space of similarities provided by bag-of-vector embeddings.

\bibliography{popa_bove_arxiv}
\bibliographystyle{acl_natbib}

\end{document}